\documentclass[runningheads]{llncs}

\usepackage[table,xcdraw,dvipsnames]{xcolor}
\usepackage{eccv}

\usepackage{eccvabbrv}

\usepackage{graphicx}
\usepackage{booktabs}
\usepackage{lipsum} 

\usepackage[accsupp]{axessibility}  

\usepackage{mathtools}

\usepackage{pifont}
\usepackage{float}
\usepackage[pagebackref,breaklinks,colorlinks]{hyperref}
\usepackage{amsmath}
\usepackage{amssymb}
\usepackage{amsfonts}

\usepackage[pagebackref,breaklinks,colorlinks]{hyperref}

\usepackage{orcidlink}

\begin{document}

\title{SPAMming Labels: Efficient Annotations \\ for the Trackers of Tomorrow} 



\author{Orcun Cetintas\inst{1, 2^\dagger} \and
Tim Meinhardt\inst{2} \and
Guillem Brasó\inst{1, 2^\dagger} \and
Laura Leal-Taix\'{e}\inst{2}}

\authorrunning{O.~Cetintas et al.}

\institute{Technical University of Munich \and NVIDIA}

\maketitle

\newcommand{\method}[1]{METHOD} 

\newcommand{\orc}[1]{{\leavevmode\color{teal}[Orcun: #1]}} 
\newcommand{\lau}[1]{{\leavevmode\color{magenta}[Kahleesi: #1]}} 
\newcommand{\gui}[1]{{\leavevmode\color{orange}[Guillem: #1]}} 
\newcommand{\moreinfo}[1]{{\leavevmode\color{gray}[More info: #1]}}
\newcommand{\tim}[1]{{\leavevmode\color{blue}[Tim: #1]}} 
\newcommand{\REF}{\textcolor{red}{\textbf{REF}}}
\newcommand{\TODO}[1]{{\leavevmode\color{brown}[TODO: #1]}} 
\newcommand{\useful}[1]{{\leavevmode\color{violet}[Useful: #1]}}

\newcommand{\modelname}{\textcolor{\colorS}{\textbf{S}}\textcolor{\colorP}{\textbf{P}}\textcolor{\colorA}{\textbf{A}}\textcolor{\colorM}{\textbf{M}}\@\xspace} 
\newcommand{\modelnameSM}{\textcolor{\colorS}{\textbf{S}}{\textbf{P}}{\textbf{A}}\textcolor{\colorM}{\textbf{M}}\@\xspace} 
\newcommand{\modelnameSPM}{\textcolor{\colorS}{\textbf{S}}\textcolor{\colorP}{\textbf{P}}{\textbf{A}}\textcolor{\colorM}{\textbf{M}}\@\xspace} 
\newcommand{\modelnamenospace}{SPAM} 
\newcommand{\cmark}{\ding{51}} 
\newcommand{\xmark}{\ding{55}} 
\newcommand{\eqdef}{\vcentcolon=}
\newcommand{\PAR}[1]{{\noindent{\textbf{#1}}}} 

\newcommand{\colorS}{orange}
\newcommand{\colorP}{YellowGreen}
\newcommand{\colorA}{Turquoise}
\newcommand{\colorM}{RubineRed}

\newcommand{\darkred}{Maroon}
\newcommand{\red}{red}
\newcommand{\lightred}{Salmon}

\definecolor{verylightblue}{HTML}{b0e0e5}
\definecolor{lightblue}{HTML}{40E0D0}
\definecolor{verydarkblue}{HTML}{1d90ff}
\definecolor{purple}{HTML}{9775dc}

\newcommand{\lightblue}{lightblue}
\newcommand{\verylightblue}{verylightblue}
\newcommand{\darkblue}{teal}
\newcommand{\verydarkblue}{verydarkblue}
\newcommand{\purple}{purple}

\begin{abstract}
Increasing the annotation efficiency of trajectory annotations from videos has the potential to enable the next generation of data-hungry tracking algorithms to thrive on large-scale datasets. Despite the importance of this task, there are currently very few works exploring how to efficiently label tracking datasets comprehensively. In this work, we introduce SPAM, a video label engine that provides high-quality labels with minimal human intervention. SPAM is built around two key insights: i) most tracking scenarios can be easily resolved. To take advantage of this, we utilize a pre-trained model to generate high-quality pseudo-labels, reserving human involvement for a smaller subset of more difficult instances; ii) handling the spatiotemporal dependencies of track annotations across time can be elegantly and efficiently formulated through graphs. Therefore, we use a unified graph formulation to address the annotation of both detections and identity association for tracks across time. Based on these insights, SPAM produces high-quality annotations with a fraction of ground truth labeling cost. We demonstrate that trackers trained on SPAM labels achieve comparable performance to those trained on human annotations while requiring only $3-20\%$ of the human labeling effort. Hence, SPAM paves the way towards highly efficient labeling of large-scale tracking datasets. Project page: \href{https://research.nvidia.com/labs/dvl/projects/spam/}{https://research.nvidia.com/labs/dvl/projects/spam}

  \keywords{Multi-object tracking \and Data labeling \and Active learning}
\end{abstract}

\makeatletter{\renewcommand*{\@makefnmark}{}
\footnotetext{{ $\dagger$ Work done during an internship at NVIDIA.}\makeatother}

\section{Introduction}
\label{sec:introduction}

Detecting and tracking objects in time is at the core of many challenging vision applications.
To address these problems, modern multi-object tracking (MOT) approaches~\cite{trackformer, zhang2021fairmot, centertrack, zeng2022motr, jde} rely on increasingly larger amounts of annotated data.
While annotating image datasets is already costly, temporal component of videos further increases the annotation difficulty and data scale requirements. This is due to the fact that redundancies between frames cause the information density to scale poorly with the amount of data, making the overall annotation task more challenging and resource-intensive.

To mitigate these issues, researchers are spending more effort in operating with limited human annotations.
Self- and semi-supervised learning (SSL) and pseudo-labeling (PL) approaches are often used to leverage unlabeled data whenever possible~\cite{lan2023vision, litrico_2023_CVPR,kirillov2023segment}, while active learning (AL) strategies aim at selecting the most informative samples to label. All these strategies are commonly applied in various image understanding tasks, such as classification~\cite{simclr, swav} or object detection~\cite{up_detr,wei2021aligning}. 
Even if video tasks do not have any lower requirement for data, there are only a handful of works that target efficient labeling of tracking data in the video domain~\cite{voigtlaender2019mots, voigtlaender2021reducing,vondrick2011video,hetero_AL_MOT_2023}.
Existing works fail to provide \textit{comprehensive} solutions to labeling by either ignoring dense temporal component of videos~\cite{hetero_AL_MOT_2023, vondrick2011video} or working under a limited single-object setup~\cite{yuan2023active}. 
An effective video labeling engine should operate across multiple objects and frames, considering the intricate dependencies between tracks inherent in the labeling problem.

Videos have unique characteristics that can be leveraged to unleash the full potential of a modern label engine. Firstly, frame-wise similarities in videos yield data redundancies. In the context of MOT, this implies that some associations between detections are relatively easy to solve~\cite{tracktor}. 
This setup provides an ideal opportunity to leverage state-of-the-art models pre-trained to pseudo-label new data at zero cost.
Secondly, the dependencies of objects across time and space cause annotations in one frame of the video to have cascading effects in other frames. For example, solving an association for one track can resolve associations for neighboring tracks. 
In order to maximize the efficiency of the annotations, we frame the labeling task within a model that takes into account such dependencies. Thus, we adopt a track-centric labeling approach rather than a frame-centric one~\cite{hetero_AL_MOT_2023}.

We present the \modelname video label engine, which chains \textcolor{\colorS}{\textbf{S}}ynthetic pre-training, \textcolor{\colorP}{\textbf{P}}seudo-labeling, and \textcolor{\colorA}{\textbf{A}}ctive learning with a graph-based \textcolor{\colorM}{\textbf{M}}odel for optimal tracking performance with minimal human effort. 
In the image domain, there are plenty of large-scale annotated datasets that a model-driven approach can leverage for pre-training~\cite{kirillov2023segment}. However, this is not the case for multi-object tracking datasets, which are typically much smaller in scale; therefore, we advocate for a \textcolor{\colorS}{\textbf{S}}ynthetic pre-training. 
We further analyze which components of a tracking method are more affected by the synthetic-to-real gap.
This first experiment gives us an important insight: labeling should be focused on detections and associations, and we can directly use the synthetically trained reID model.
With our synthetic pre-trained model, we then propose a model-driven approach to produce \textcolor{\colorP}{\textbf{P}}seudo-labels on real data and retrain the model with its own pseudo-labels. We show that with this recipe we can even reach a performance comparable to ground truth without ever training our labeler on manually annotated data.

Synthetic pre-training and pseudo-labeling allow us to handle easy decisions and defer only uncertain or complex decisions to an annotator. 
We propose an \textcolor{\colorA}{\textbf{A}}ctive learning strategy where the labeler makes decisions at the track-level instead of frame-level, thereby utilizing the annotation budget more efficiently. Towards this goal, we solve the labeling problem with a \textcolor{\colorM}{\textbf{M}}odel, based on graph hierarchies~\cite{sushi}. This model captures long-range spatio-temporal dependencies, which means our annotations will have an impact across frames, and not only on the frame where the annotation actually happens. 
We extend the model with a detection filtering layer so that we can naturally label detection and association within the same framework.
In summary, we propose \modelname, a video label engine that significantly reduces annotation cost by dealing with easy scenarios with a strong MOT model pre-trained on (i) \textcolor{\colorS}{\textbf{S}}ynthetic data, followed by (ii) \textcolor{\colorP}{\textbf{P}}seudo-labeling real data for its own re-training. Harder detection and association decisions are carefully selected by an (iii) \textcolor{\colorA}{\textbf{A}}ctive learning scheme that works on a (iv) \textcolor{\colorM}{\textbf{M}}odel based on graph hierarchies. This allows us to efficiently use the annotator's budget and account for spatio-temporal dependencies in the labels. 
The resulting model generates labels comparable to ground truth performance, requiring only $3-20\%$ of manual annotation effort across three diverse datasets.

\section{Related Work}
\label{sec:related_work}


\PAR{Pre-training for tracking}.
Over the years, the multi-object tracking (MOT) community has annotated many challenging benchmark datasets~\cite{motchaijcv, mot20, dancetrack}.
However, due to the significant costs associated with manually labeling temporal data, most benchmarks only provide few annotated video sequences.
To obtain top performance, modern data-hungry tracking methods are usually pre-trained on additional video and tracking datasets~\cite{4587581, zhang2017citypersons, dollar2009pedestrian, xiao2017joint, zheng2017person}.
Many works also rely on relevant image recognition~\cite{shao2018crowdhuman} datasets to boost their per-frame object detection performance.
An alternative pre-training direction avoids human annotations entirely by relying on simulated synthetic data~\cite{gaidon2016virtual,fabbri2018learning,sun20eccv, Hu3DT19,motsynth}.

Our label engine also pre-trains the labeling model synthetically, minimizing its total annotation footprint.
To this end, we demonstrate the effectiveness of synthetic pre-training~\cite{motsynth} for generating pseudo-labels for self-supervision.

\PAR{Self-supervision in the temporal domain}.
To benefit not only from already annotated datasets, many works explore the utilization of available unlabeled data sources, \ie, how to perform self-supervised learning for videos.
Notable examples leverage cycle-consistency-based correspondences~\cite{CVPR2019_CycleTime, jabri2020space}, color propagation~\cite{vondrick2018tracking} or augmentation-based consistencies~\cite{bastani2021self, liu2023uncertainty}.
Furthermore, \cite{meng2023tracking} uses contrastive learning over re-identification (reID) embeddings to perform unsupervised association learning.
More recently, several works~\cite{lan2023vision, litrico_2023_CVPR,kirillov2023segment, Kaul22} in the image domain demonstrate the effectiveness of training on pseudo-labels generated by strong underlying labeling models.
In the video domain, this type of self-supervision is mostly unexplored.

We demonstrate the effectiveness of pseudo-labels for the temporal domain by improving the label quality of our SPAM labeler via self-supervision on its own output.
The necessary signal-to-noise ratio for this training loop is obtained by pre-training a strong graph-based model on synthetic data.

\PAR{Efficient labeling of video datasets}.
While significant effort has been devoted towards increasing the labeling efficiency of image datasets~\cite{AcunaCVPR18, kirillov2023segment}, similar approaches for the video and tracking domain are still relatively limited.
Earlier works reduce the number of frames to annotate within individual videos but still rely heavily on human annotators~\cite{vondrick2010efficiently}.
Other solutions are limited to only a single object per sequence~\cite{voigtlaender2021reducing, dai2021video} or still rely on human bounding box annotations~\cite{voigtlaender2019mots} to label segmentation masks across time.
The authors of~\cite{hetero_AL_MOT_2023} select key frames based on their expected impact on tracking performance.
While effective within its scope, this approach only yields sparse frame annotations, which are insufficient for training all modern trackers.
To reduce the human annotation effort for trajectories,~\cite{manen2017pathtrack} proposes an efficient video interface that still requires labeling all trajectories.

In contrast, our label engine performs active learning for multiple objects across all video frames and is the first of its kind to tackle long-term tracking problems via a hierarchical labeling approach.
The uncertainty measures provided by our graph-based model allow us to focus expensive annotation efforts on individual tracks instead of labeling entire frames in a brute-force manner.

\PAR{Multi-object tracking models.}
Tracking objects across time in a video is an established problem with a long research history. 
The most dominant tracking-by-detection (TbD) paradigm splits the task into two steps: (i) detecting objects in every frame and (ii) associating them to object trajectories, also referred to as data association.
Modern trackers have explored performing (i) and (ii) jointly~\cite{dttd,tracktor,trackformer,centertrack}, or providing solutions for (ii) relying on motion models~\cite{motiontrack,6130233, Alahi_2016_CVPR, Robicquet2016LearningSE,5995311,scovanner2009learning, 5459260, 5995468,bytetrack}, appearance models~\cite{ghost,zhang2021fairmot,qdtrack,utm, Leal-Taixe_2016_CVPR_Workshops, Ristani_2018_CVPR,chen2018real,famnet} or more complex techniques involving transformers~\cite{zeng2022motr,cai2022memot}.
An alternative direction of work relies on graph formulations~\cite{berclaz2011multiple, 4270205, 5995604, network_flows_tracking,Leal-Taixe_2014_CVPR} to utilize the TbD paradigm. 
In this formulation, nodes model object detections, and edges, association hypotheses.
This long-studied formulation has recently been revived with remarkable success with the help of learnable graph neural network (GNN)architectures~\cite{mpntrack, MPNTrackSeg, 9561110,sushi}.

For our label engine to generate long-term pseudo-labels and uncertainty estimations, we rely on a strong graph-based tracker~\cite{sushi}.
To incorporate both detection and association labeling within a unified graph framework, we introduce a novel hierarchy level dedicated to detection filtering.

\section{SPAM}
\label{sec:method}

\subsection{Overview}

\begin{figure}[tb]
  \centering
  \includegraphics[width=\columnwidth]{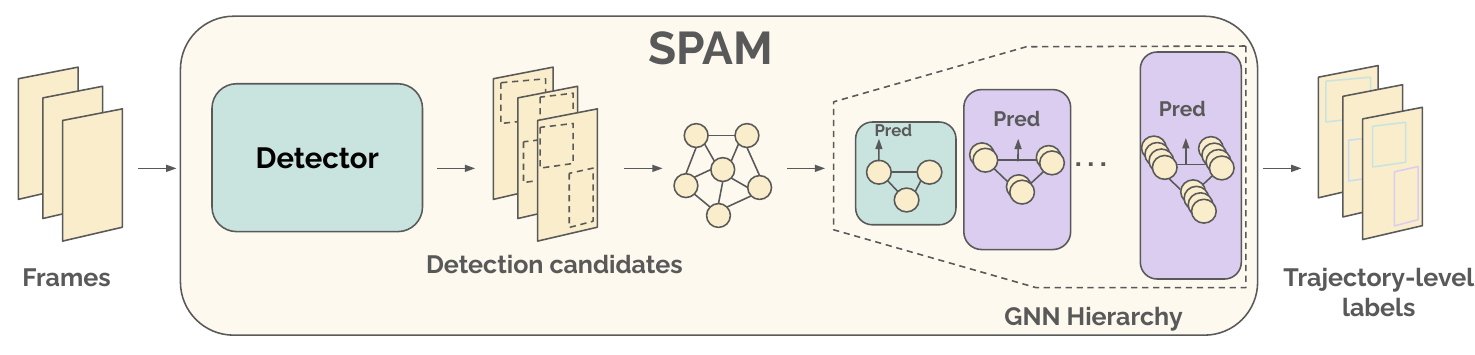}
  \caption{Overview of the SPAM model. We first generate a set of detection candidates with our detector. Hierarchical GNNs then classify these candidates into valid and invalid objects via node classification, and assign identities through edge classification.}
  \label{fig:pipeline}
\end{figure}


Our main goal is to label multi-object tracking datasets \textit{with minimal annotation effort while enabling optimal tracking performance}. Our key observation is that most association cases can be solved easily by models pre-trained on synthetic data. 
Therefore, no human effort is needed, and we can instead use our model's predictions as pseudo-labels. 
Decisions that are more complex or uncertain are selected by an active learning scheme to be annotated manually. 

In order to perform pseudo-labeling and active learning, we require a model that can capture long-term spatio-temporal dependencies between tracks.
We leverage a hierarchical graph neural network formulation and cast the annotation task over a graph. 
Our architecture~(\cref{fig:pipeline}) draws inspiration from SUSHI~\cite{sushi}, which demonstrated remarkable association performance. 
Our model's scope, however, goes beyond association. Given the constraints of the labeling task, we need to equip our model with the ability to reason about detections before performing association. To address this broader goal, our GNN hierarchy, also includes the capability to accept or reject detections by classifying nodes into correct and incorrect object hypotheses. 

Our overall model is pre-trained on synthetic data~\cite{motsynth} and further fine-tuned on its own predicted pseudo-labels over the target dataset that we aim to annotate. This recipe allows for strong labelling performance \textit{without any human annotations}, and provides the basis for a scalable video label engine. After this training loop, we use our updated model's predictions as pseudo-labels to annotate a large portion of the target dataset. Decisions that are more complex or uncertain are selected by an active learning scheme to be annotated manually. The resulting set of labels achieves near-GT performance with significantly fewer manual annotations, thereby vastly enhancing the efficiency of the overall process. We provide an overview of our training and annotating recipe in \cref{fig:SPAM}.

\subsection{Graph Hierarchies as a Model for Labeling}
\label{section:graphlabeling}

\PAR{Problem formulation.} Our model builds upon the standard graph formulation~\cite{network_flows_tracking} where, given a set of object candidates $\mathcal{O}$, our goal is to obtain a set of valid objects $\mathcal{O}_v \subset \mathcal{O}$ and their corresponding trajectories $\mathcal{T}$. Each trajectory $T_k \in \mathcal{T}$ consists of a set of objects sharing the same identity or, equivalently, the set of edges connecting those nodes. Specifically, we model each object candidate $o_i \in \mathcal{O}$ as a node and each association hypothesis among candidates as an edge in an undirected graph $G=(V, E)$. Within this formulation, nodes $u \in V$ could be classified into valid objects if $u \in \mathcal{O}_v$. Analogously, edges can be classified into correct or incorrect hypotheses depending on whether they belong to some trajectory or not. This formulation can be generalized to a hierarchical setting by splitting video sequences into non-overlapping subsequences, \ie, disjoint subgraphs, and progressively merging them~\cite{sushi}. 

\PAR{Background on graph hierarchies for MOT.} Our model consists of a hierarchy of GNNs operating over the graph formulation described above. Intuitively, starting from detection candidates, at each hierarchy level, GNN hierarchy subsequently merge tracklets from the previous level into longer ones. Nodes and edges are represented by initial embeddings and message passing is used to share information within the graph in order to update node and edge embeddings with richer information. The GNNs are trained to classify edges into active and inactive association hypotheses and they share the same features, network architecture and learnable weights across hierarchy levels. In addition to the edge features reported in ~\cite{sushi}, for each object candidate $o_i$ we concatenate its timestamp $t_i$, normalized bounding box coordinates and dimensions $(x_i, y_i, h_i, w_i)$ and corresponding confidence from the detector $c_i$ and embed it as node features.

\begin{figure}[tb]
  \centering
  \includegraphics[width=\columnwidth]{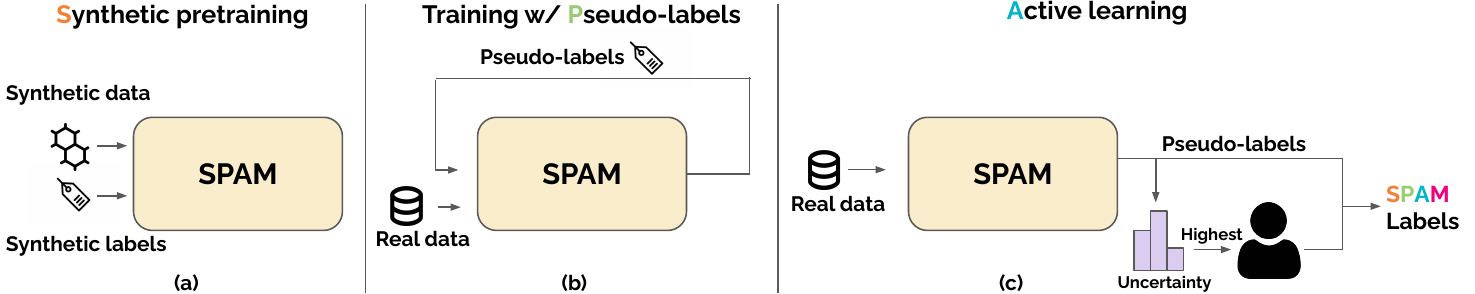}
  \caption{Overview of the SPAM training and annotation pipeline. (a) Initial model training on synthetic data. (b) Application of SPAM to generate pseudo-labels without incurring manual annotation costs on a real dataset, followed by self-training on pseudo-labels. (c) Real dataset labeling using pseudo-labels and an uncertainty-based active learning approach.}
  \label{fig:SPAM}
\end{figure}
\PAR{Classifying nodes.}  We aim to minimize the annotation effort; therefore, we ask the annotator to operate on a fixed graph of detection candidates, \ie annotator does not label new detections after the initial set given by detector. This is why we aim to create an initial graph that is over-complete. In order to do so, we use a low confidence threshold for our detector and obtain a large number of detection candidates. While this ensures a high detection recall, it also introduces a larger number of false positive detections that our model needs to address.
As we show in experiments, the original SUSHI architecture~\cite{sushi} cannot handle such false positives, which is why we propose a new layer in the hierarchy, $GNN_{node}$. 
Intuitively, $GNN_{node}$  leverages temporal information to validate object candidates. We use its output node embeddings and feed them to a lightweight classifier to enable false positive detection filtering. By doing so, we can consider a large number of input proposals achieving high detection recall by leveraging the spatio-temporal reasoning ability of our GNN for detection candidates.

\PAR{Pre-training and \textit{annotation-free} fine-tuning.} We pre-train all of our model components on the large-scale synthetic dataset MOTSynth~\cite{motsynth}.  To boost our model's performance and bridge the domain gap with our target dataset for annotation, we conduct further fine-tuning on the target dataset after pre-training. We achieve this by utilizing the raw predictions of our model on the target dataset and employing them as pseudo-labels for \textit{self-training}. As we show in our experimental section, this strategy yields a significant performance improvement without incurring any human annotation cost. 

\subsection{How to Label on Graph Hierarchies?}\label{sec:howtolabel}

\PAR{Manual labeling.} Labeling video data at the box level involves several steps: identifying objects of interest, localizing their boundaries with bounding boxes, and assigning them unique identities. Typically, annotators achieve this by examining each frame, detecting and localizing objects individually, and then assigning identities based on temporal context from neighboring frames. In terms of annotation effort in clicks, \textit{detecting} and \textit{localizing} objects typically require two clicks (one for the top-left corner and one for the bottom-right corner of the bounding box), while \textit{associating} objects across time requires one click (identifying the next occurrence of the object in subsequent frames), as reported in ~\cite{motchaijcv, mot20}. This manual process is labor-intensive, costly, and severely limits the availability of annotated video data.
\begin{figure}[tb]
  \centering
  \includegraphics[width=0.85\columnwidth]{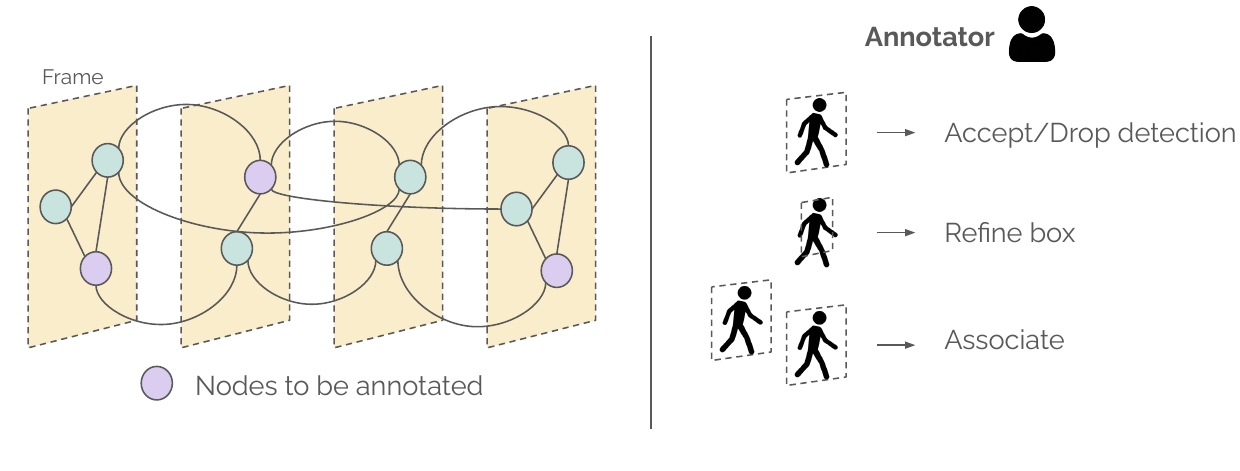}
  \caption{Our graph-based labeling pipeline begins with the selection of nodes for annotation. For each node to be annotated, the annotator could be asked to validate the \textit{detection}, improve the \textit{localization} by refining the box or perform \textit{association}.}
  \label{fig:graphann}
\end{figure}

\PAR{Graph-based labeling.} Utilizing a graph formulation for detection and tracking enables explicit modeling of objects in the scene and their interactions. In this framework, each node of the graph represents an object, allowing a manual annotation of an object to be framed as selecting a node on this graph. Given a set of nodes to be annotated, possible annotator decisions are: (i) accepting/rejecting a \textit{detection} as we are working on a large set of detection candidates (one click), (ii) refining the bounding-box to improve \textit{localization} (two clicks, one for the top-left corner and one for the bottom-right corner of the bounding box) and (iii) \textit{associating} a node across time (one click) as shown in  \cref{fig:graphann}.

\PAR{Annotation acquisition.} The core idea of our approach is to use our model decisions to guide the annotation process. For the vast majority of scenarios, we  directly use our model's predictions as pseudo-labels. For a small subset of more challenging decisions where our model exhibits uncertainty, we validate our decisions with an annotator as described above. Within our graphs, we adopt an uncertainty-based approach to identify which nodes require annotations. Specifically, for each node $v \in V$, we compute its uncertainty as $\text{uncert}(v) \eqdef \text{max}_{u \in \text{N}_v} H({\hat{y}_{(v, u)}})$, where $\text{N}_v$ denotes the set of neighboring nodes of $v$ and $\hat{y}_{(v, u)}$ is the edge prediction for edge $(v, u)$:

\begin{equation}
    H({\hat{y}_{(v, u)}}) \eqdef - (\hat{y}_{(v, u)} \log \hat{y}_{(v, u)} + (1-\hat{y}_{(v, u)}) \log(1-\hat{y}_{(v, u)}))
\end{equation}

At the $GNN_{node}$ level, where we have node predictions instead of edge predictions, we simply compute $\text{uncert}(v)$ as $H({\hat{y}_{(v)}})$ instead.
Subsequently, we refer the nodes with the highest uncertainty scores to the annotator for manual annotation and use the remaining predictions as pseudo-labels for the rest of the nodes.

\PAR{Hierarchical label generation.} To fully leverage our hierarchical graph formulation, we apply our label acquisition approach described above in a hierarchical manner. Intuitively, we distribute our annotation budget $B$ across $L$ hierarchical levels as $B_1, \dots, B_L$, with $B_1 +\dots + B_L = B$.
In deeper hierarchy levels, nodes represent tracklets instead of single detections. This allows us to propagate annotator decisions for the entire cluster instead of individual detections, resulting in solving multiple uncertainties with a single annotation. 
As a result, we utilize the annotation budget more effectively while also mitigating error propagation within our pseudo-labels.

\section{Experiments}
\label{sec:experiments}

\begin{figure}[tb]
  \centering
  \includegraphics[width=0.72\columnwidth]{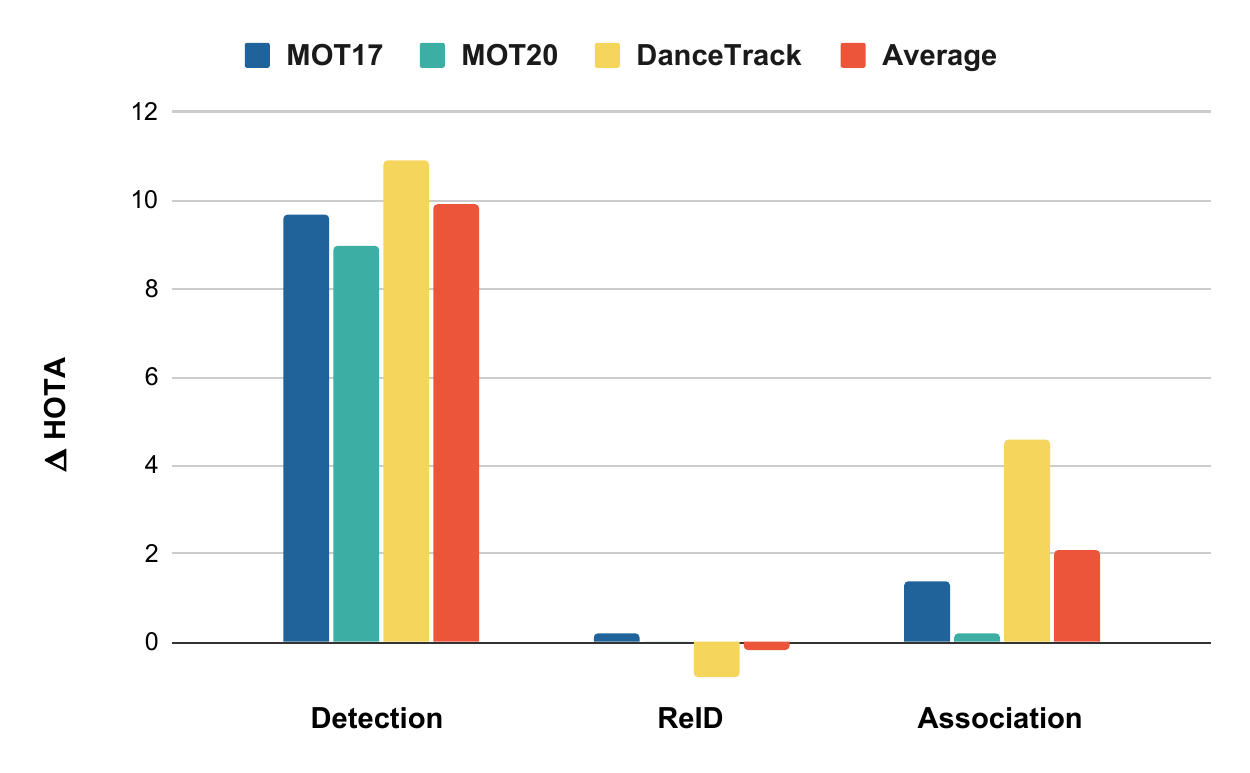}
  \caption{Analysis of performance gap between training a model on synthetic and real data for the three most common tracking components: detection, association, re-identification.}
  \vspace{-0.3cm}
  \label{fig:experiment_synth}
\end{figure}

\subsection{Datasets and Metrics}
\PAR{Datasets.} We conduct experiments on three public benchmarks:

\begin{itemize}
    \item \textbf{MOT17}~\cite{motchaijcv} is an established benchmark containing 14 challenging video sequences, exhibiting diverse characteristics, including variations in camera motion, viewpoint, and pedestrian densities.

    \item \textbf{MOT20}~\cite{mot20} contains a total of 8 video sequences, all featuring extremely crowded scenes. Due to their high pedestrian density, the emphasis is on addressing highly challenging scenarios characterized by frequent and prolonged occlusions.

    \item \textbf{DanceTrack}~\cite{dancetrack} contains 100 video sequences. These sequences feature dancing individuals with similar appearances, facilitating the evaluation of tracker robustness against highly challenging motion patterns where appearance information has a relatively minor influence. This non-standard setup often leads to a performance drop in trackers~\cite{dancetrack}.
\end{itemize}

\PAR{Metrics.}
We assess the performance of trackers using three established metrics:

\begin{itemize}
    \item \textbf{MOTA}~\cite{clear} assesses trajectory coverage, by combining object detection measurements with identity switches, with the latter playing a relatively minor role. 
    \item \textbf{IDF1}~\cite{IDF1} focusses on identity preservation over the sequences and provides a measure of association performance.
    \item \textbf{HOTA}~\cite{hota} is a recently introduced metric designed to strike a balance between association and object detection performance within a single score.
\end{itemize}

\subsection{Implementation Details}
\PAR{Architecture and training.} Our model consists of three main components: (i) a detector network, based on YOLOX~\cite{ge2021yolox}, (ii) a ResNet-based ReID network~\cite{he2016deep}, and (iii) a GNN hierarchy, all pretrained on the synthetic sequences of MOTSynth~\cite{motsynth}. After pretraining, we use the resulting model to obtain pseudo-labels for our target MOT dataset, and then fine-tune our detector and GNN on them. Note that our \textcolor{\colorM}{M}odel is \textit{never trained on human annotations}, it is only trained on synthetic annotations and its own pseudo-labels.

\PAR{Experimental setup.}  Online trackers are real-time tools suitable for deployment on robots or autonomous vehicles, whereas offline trackers are better suited for labeling tasks. Therefore, we structure our experiments to utilize \modelname for labeling and use these labels with online trackers. To evaluate the quality of \modelname labels, we train two state-of-the-art online trackers with these labels: ByteTrack~\cite{bytetrack} and GHOST~\cite{ghost}.
For validation, we follow the standard half-split setting~\cite{zhang2021fairmot, bytetrack, ghost} over MOT17 sequences. Annotation budgets are computed based on number of clicks as described in \cref{sec:howtolabel}.

\subsection{\textcolor{\colorS}{S}ynthetic Pretaining} 

Several works have shown that synthetic pre-training is a strong starting point for multi-object tracking methods~\cite{bytetrack, motsynth}. In this section, we aim to analyze the gap between training a model on synthetic data and training it with real data for the three most common tracking components: detection, association, re-identification.
For each setting, only the specific component that we are analysing is trained on synthetic data while other components are all trained on real data. We perform this experiment on three datasets and plot the relative HOTA difference (a positive score indicates superior performance with real data training) when training on real data vs. synthetic data in \cref{fig:experiment_synth}. 
Our first observation, is that the detector is the component that suffers the domain gap the most, with a performance drop of 9.9 HOTA points. Re-identification, on the other hand, has a similar performance when trained on synthetic or real data.
Lastly, association performance is 2.1 HOTA higher with real data training. 
This first experiment gives us an important insight: labeling and retraining should be focused on detection and association components, and we can directly use the synthetically trained reID model.

\subsection{Towards a \textcolor{\colorM}{M}odel for Labeling}

\begin{table}[tb]
\center
\tabcolsep=0.11cm

    \resizebox{\columnwidth}{!}{
    \begin{tabular}{l | c c | c | c c c c}
     \toprule
     Model & High Conf. Box & Low Conf. Box & $GNN_{node}$ & MOTA $\uparrow$& IDF1 $\uparrow$ & HOTA $\uparrow$ & DetA $\uparrow$ \\ [0.5ex] 
     \midrule
     A & \cmark & \xmark & \xmark & 64.4 & 74.7 & 59.9 & 55.9 \\
     B & \cmark & \cmark & \xmark & 60.6 & 71.4 & 58.5  & 54.9 \\
     C & \cmark & \cmark & \cmark & \colorbox{green!30}{\textbf{65.4}} & \colorbox{green!30}{\textbf{75.1}} & \colorbox{green!30}{\textbf{60.4}} & \colorbox{green!30}{\textbf{56.6}} \\
     \midrule
    \end{tabular}}

\caption{Ablation study investigating the impact of incorporating low confidence bounding boxes with and without the $GNN_{node}$.}
\vspace{-0.7cm}
\label{table:nodegnn}
\end{table}

\PAR{Graph-based validation of detection candidates.}
In our framework, annotators have the capability to eliminate detections, but we refrain from requesting them to annotate new detections in order to maintain minimal annotation effort. Instead we start from an initial graph formed by an over-complete set of detections. Towards this end, we use a low confidence threshold for our detector, thereby reducing the number of false negatives in our initial graph and ensuring a high recall. 
As a result, this leads to an increase in the number of false positives incorporated into the graph, a tendency that is further accentuated by the fact that our detector is exclusively trained on synthetic data.
We propose to leverage the graph structure itself to validate the detections based on temporal consistency. Specifically, we add an additional GNN-layer in our hierarchy, $GNN_{node}$, which intuitively  examines the spatial and temporal consistency of the bounding box candidates and filters out implausible ones.
In ~\cref{table:nodegnn}, we assess the effect of this design choice.
Model A uses only high confidence bounding boxes, model B creates a graph with also low confidence bounding boxes but without $GNN_{node}$, while C includes $GNN_{node}$.
As we can see, simply adding low confidence boxes results in a drastic drop in performance on all metrics. Only by adding our $GNN_{node}$ we are able to take full advantage of the low confidence boxes without penalizing our performance with too many errors. 
%

\begin{table}[tb]
\center
\tabcolsep=0.11cm
    \resizebox{\columnwidth}{!}{
    \begin{tabular}{l c c c |  c c c | c c c }
     \toprule
     Method  & HOTA $\uparrow$ & MOTA $\uparrow$  & IDF1 $\uparrow$ & HOTA $\uparrow$ & MOTA $\uparrow$  & IDF1 $\uparrow$ & HOTA $\uparrow$  & MOTA $\uparrow$ & IDF1 $\uparrow$ \\ [0.5ex] 
     \midrule
    \multicolumn{1}{l}{} & \multicolumn{3}{c}{MOT17} &   \multicolumn{3}{c}{MOT20} &   \multicolumn{3}{c}{DanceTrack} \\
     \midrule
     MeMOT~\cite{cai2022memot} & 56.9 & 72.5 & 69.0 & 54.1  & 66.1 & 63.7 & -- & -- & --  \\
     TrackFormer~\cite{trackformer} & 57.3 & 74.1 & 68.0   & 54.7 & 65.7 &  68.6 & -- & -- & -- \\
     MOTR~\cite{zeng2022motr} & 57.8 & 68.6 & 73.4 & -- & -- & -- & 54.2  &  51.5  & 79.7\\
     FairMOT~\cite{zhang2021fairmot} & 59.3 & 73.7  & 72.3 & 54.6  & 61.8  & 67.3 & 39.7 & 82.2 & 40.8  \\
     QDTrack~\cite{qdtrack} & 63.5 & 78.7 & 77.5 & 60.0 & 74.7 & 73.8 & 54.2 & 87.7  & 50.4  \\
     ByteTrack$^{\dagger}$~\cite{bytetrack} & 62.8 & 78.9 & 77.1 & 60.4 &  74.2 & 74.5   & 47.7 & \cellcolor[HTML]{9AFF99} \textbf{89.6}& 53.9     \\
     GHOST~\cite{ghost} & 62.8 & 78.7 & 77.1 & 61.2 & 73.7 & 75.2 & 56.7 & 91.3 & 57.7 \\
    UTM~\cite{utm} & 64.0 & \cellcolor[HTML]{9AFF99} \textbf{81.8} & 78.7 & 62.5 & \cellcolor[HTML]{9AFF99} \textbf{78.2} & 76.9 & -- & -- & -- \\
    MotionTrack~\cite{motiontrack} & 65.1 & 81.1 & 80.1 & 62.8 & 78.0 & 76.5 & -- & -- & -- \\
     SUSHI~\cite{sushi} & 66.5 & 81.1 & 83.1 & 64.3 & 74.3  & 79.8  & 63.3& 88.7  & \cellcolor[HTML]{9AFF99} \textbf{63.4}   \\
     \midrule
     \textbf{\modelnameSM (Ours)} & \cellcolor[HTML]{9AFF99} \textbf{67.5} & 80.7 & \cellcolor[HTML]{9AFF99} \textbf{84.6} & \cellcolor[HTML]{9AFF99} \textbf{65.8} & 76.5 &  \cellcolor[HTML]{9AFF99} \textbf{81.9} & \cellcolor[HTML]{9AFF99} \textbf{64.0} &  89.2 & \cellcolor[HTML]{9AFF99} \textbf{63.4}  \\
     
    \midrule

    \end{tabular}}

\caption{Test set results on MOT17, MOT20 and DanceTrack when using \modelnameSM solely as a tracker.}
\vspace{-0.8cm}
\label{table:alltracking}
\end{table}

\PAR{\textcolor{\colorS}{S}PA\textcolor{\colorM}{M} is a SOTA tracker.} Before moving to labeling with our model, we want to verify its overall performance and generalization capabilities. Only by using a state-of-the-art model we will be able to reduce annotation effort as much as possible.
We therefore pre-train our model on the synthetic dataset, and finetune it on the specific real dataset using groundtruth.
For a fair comparison, we obtain object detections from a YOLOX detector~\cite{ge2021yolox} trained following~\cite{bytetrack, ghost, sushi}.
Our model outperforms state-of-the-art by 1.0 HOTA point on MOT17, 1.5 HOTA points on MOT20, and 0.7 HOTA points on DanceTrack as shown in \cref{table:alltracking}. Furthermore, on MOT17 and MOT20, there is also a significant improvement in IDF1, showing the identity-preserving capabilities of the model, which is especially important for our video annotation setting.
Even if being state-of-the-art is not the main goal of this paper, having a strong performing model allows us to have an excellent starting point for producing pseudo-labels.
We would like to emphasize that, we use ground truth annotations only for this experiment just to compare our model fairly with other trackers. From this point onward, we will not use ground truth labels. We will use our synthetic pretraining weights and focus on the pseudo-labeling and active learning aspects of our method to show how to annotate multi-object tracking datasets.

\subsection{Enhancing Annotations with \textcolor{\colorP}{P}seudo-labels and \textcolor{\colorA}{A}ctive Learning}

\PAR{Retraining our \textcolor{\colorM}{M}odel on \textcolor{\colorP}{P}seudo-labels.} In this experiment, we compare two models. The first one, \textbf{\textcolor{\colorS}{S}PA\textcolor{\colorM}{M}}, is only trained on synthetic data. The second one, \textbf{\textcolor{\colorS}{S}\textcolor{\colorP}{P}A\textcolor{\colorM}{M}}
, is trained in three stages: (i) pre-training on synthetic data, (ii) pseudo-labeling the real dataset with our model, and (iii) re-training our model on pseudo-labels. 
In ~\cref{table:pseudo}, we observe a significant improvement when using pseudo-labels to re-train our model: an increase of 4-6 HOTA, 4-6 MOTA, and 4-7 IDF1 points on MOT17, MOT20, and DanceTrack, respectively. Note that these improvements come at zero human annotation effort as this process requires no manual annotations. 
%

\noindent 
\begin{figure}[t]
\begin{minipage}{0.44\textwidth}
  \centering
  \label{table:pseudo}
  \tabcolsep=0.11cm
  \resizebox{\textwidth}{!}{%
    \begin{tabular}{l | c | c c c}
            \toprule
     Model   & Pseudo-label & HOTA $\uparrow$ & MOTA $\uparrow$& IDF1 $\uparrow$   \\ [0.5ex] 
     \midrule
     \multicolumn{5}{c}{MOT17} \\
     \midrule
     \modelnameSM & \xmark & 60.0 & 65.3 & 75.1  \\
     \modelnameSPM & \cmark & \colorbox{green!30}{\textbf{63.8}} & \colorbox{green!30}{\textbf{69.2}} & \colorbox{green!30}{\textbf{79.7}}   \\
     
     \midrule
     \multicolumn{5}{c}{MOT20} \\
     \midrule 
     \modelnameSM & \xmark & 52.2   & 65.4 & 70.3 \\
     \modelnameSPM & \cmark & \colorbox{green!30}{\textbf{58.7}} & \colorbox{green!30}{\textbf{70.2}} & \colorbox{green!30}{\textbf{77.1}} \\
     \midrule
     
     \multicolumn{5}{c}{DanceTrack} \\
     \midrule
     \modelnameSM & \xmark & 41.8  & 65.2 & 43.4 \\
     \modelnameSPM & \cmark  & \colorbox{green!30}{\textbf{48.1}} & \colorbox{green!30}{\textbf{71.0}} & \colorbox{green!30}{\textbf{50.2}} \\
     \midrule

      \end{tabular}

    }
    \captionof{table}{Performance boost obtained by our model when retraining with its own pseudo-labels incurring no manual annotation cost.}
   \end{minipage}%
\hfill
\begin{minipage}{0.52\textwidth}
  \centering
  \vspace{-1cm}
  \includegraphics[width=1.05\textwidth]{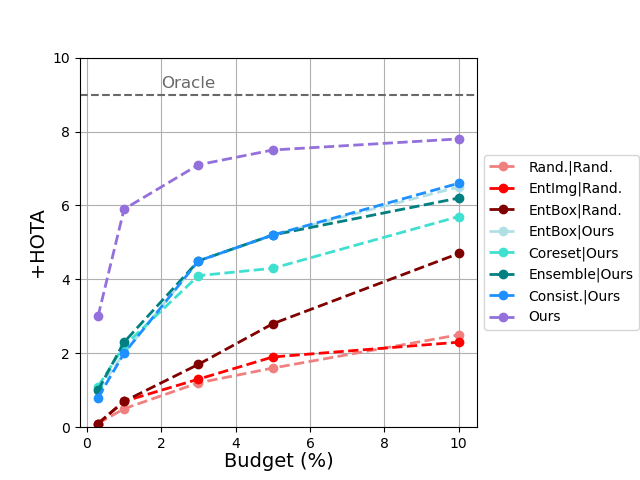}
  \vspace{-0.77cm}
  \captionof{figure}{Impact of various active learning strategies on the quality of labeling.}
  \label{fig:experiment_al}
\end{minipage}
\end{figure}

\PAR{\textcolor{\colorA}{A}ctive learning on graph hierarchies.}
We now test the effectiveness of our proposed active learning labeling on graph hierarchies, and show a comprehensive comparison with several baselines in~\cref{fig:experiment_al}.
We directly measure the accuracy of the output labels given an increasing percentage of manual annotations selected by different acquisition functions on MOT17.
The vertical axis shows the increase in HOTA over the baseline performance, \ie, the performance without active learning. \textit{Oracle} is our upper bound and shows the performance if all possible labels are set to their correct value by the annotator.
Every method has two different labeling schemes: one for detections (nodes), and one for associations (edges) as explained in \cref{sec:howtolabel}. In addition to our method, we explore active learning strategies commonly used in object detection, including Coreset~\cite{coreset}, Ensemble~\cite{ensemble}, and Consistency over augmented inputs~\cite{al_ili}. The naming is based on the active learning strategy for detection in the first part and the strategy for association in the second part. For instance, \textit{EntImg|Rand.} indicates image level entropy-based selection for detection and random selection for association.

The methods shown in red shades and circle markers compare three detection labeling methods while keeping association label selection random. \textcolor{\darkred}{Box-level labeling} is far superior to commonly-used \textcolor{\red}{image-level labeling},  or \textcolor{\lightred}{random labeling}. This highlights the potential of performing active learning at the object level rather than the frame level.
The methods in blue shades and triangle markers all use our proposed active learning selection on graph hierarchies for the association, and compare several detection selection schemes.
The \textcolor{\lightblue}{Coreset}~\cite{coreset}, which prioritizes diversity of the data over uncertainty, emerges as the poorest performing approach. This observation underlines that, when accurate predictions are available, an uncertainty-based approach proves to be a more effective complement to pseudo-labels. \textcolor{\verylightblue}{Box-level entropy}, \textcolor{\darkblue}{ensembles}, and \textcolor{\verydarkblue}{consistency} all perform similarly.
Finally, \textcolor{\purple}{our method}, with both detection and association label selection on graph hierarchies is superior to all compared baselines, staying at only 1.2 points from Oracle performance. Notably, we also unify the AL strategy for detection and association with our approach. 

\subsection{Putting Everything Together: \modelname Labels in Action}\label{sec:together}
\begin{table}[tb]
\center
\tabcolsep=0.11cm

    \resizebox{\columnwidth}{!}{
    \begin{tabular}{l | l | c | c | c | c c c}
     \toprule
     Model & Labels & Pseudo-labels & Active Learn. & Ann. Effort $\downarrow$ & HOTA $\uparrow$ & MOTA $\uparrow$ & IDF1 $\uparrow$\\ [0.5ex] 
     \midrule
     ByteTrack~\cite{bytetrack}& Ground Truth & \multicolumn{2}{|c|}{--} & 100\% & \colorbox{green!30}{\textbf{52.6}} & 60.4 & 65.7\\
     \midrule 
     ByteTrack~\cite{bytetrack}&\modelnameSM & \xmark & \xmark  & \xmark & 49.6 & 57.8 & 63.1\\
     ByteTrack~\cite{bytetrack}&\modelnameSPM & \cmark & \xmark & \xmark & 50.5 & 59.5 & 64.5\\
     ByteTrack~\cite{bytetrack}&\modelname & \cmark & \cmark & 3.3\%  & 52.5 & \colorbox{green!30}{\textbf{61.8}} & \colorbox{green!30}{\textbf{66.2}} \\
    \midrule
    \end{tabular}}

\caption{Effect of (i)  \textcolor{\colorS}{S}ynthetic training, (ii) Retraining our model with our \textcolor{\colorP}{P}seudo-labels, and (iii) \textcolor{\colorA}{A}ctive learning-based acquisition and annotation of a small percentage of real data, on our label quality on the MOT17 validation set.}

\label{table:val}
\vspace{-0.5cm}
\end{table}

\PAR{Training trackers with our labels.} In this experiment, we ablate the effects of our \modelname pipeline on the actual usecase of training a model with our provided labels. We ablate label quality under the following settings: (i) \textcolor{\colorS}{S}ynthetic training, (ii) Retraining our models on our \textcolor{\colorP}{P}seudo-labels, (iii) \textcolor{\colorA}{A}ctive learning based acquisiton and annotation of a small percentage of real data. 
We train state-of-the-art online tracker ByteTrack~\cite{bytetrack} from scratch with the aforementioned sets of labels and report its performance on the MOT17 validation set in ~\cref{table:val}. 
The goal performance we want to reach is the one achieved by the model trained on groundtruth annotations (row 1). Row 2 shows our model trained only on synthetic data, which is 3 HOTA points below our goal. Using pseudo-label based training of our model increases performance by almost a HOTA point, and the ground truth level performance is achieved by also using our active learning method to select which samples to label. By annotating only $3.3\%$ of the data, we can achieve the same results as the model trained on the ground truth data with $100\%$ annotation effort. This shows the potential of our \modelname in reducing annotation effort for multi-object tracking datasets.

\begin{table}[tb]
\center
\tabcolsep=0.11cm

    \resizebox{0.8\columnwidth}{!}{
    \begin{tabular}{l | c | c | c c c}
     \toprule
     Method & Labels & Ann. Effort $\downarrow$ & HOTA $\uparrow$ & MOTA $\uparrow$ & IDF1 $\uparrow$\\ [0.5ex] 
     \midrule
     \multicolumn{6}{c}{MOT17 Test} \\
     \midrule
     ByteTrack~\cite{bytetrack} & GT & 100\% & 50.6 & 60.4 & 61.1 \\
     ByteTrack~\cite{bytetrack} & \modelname & 3.3\%  & \colorbox{green!30}{\textbf{51.6}} & \colorbox{green!30}{\textbf{64.0}} &\colorbox{green!30}{\textbf{63.0}} \\
     \midrule
     GHOST~\cite{ghost} & GT & 100\% & 49.5 & 58.0 & 59.0 \\
     GHOST~\cite{ghost} & \modelname & 3.3\%  & \colorbox{green!30}{\textbf{51.3}} & \colorbox{green!30}{\textbf{61.9}} & \colorbox{green!30}{\textbf{62.1}} \\

     \midrule
     \multicolumn{6}{c}{MOT20 Test} \\
     \midrule
     ByteTrack~\cite{bytetrack} & GT & 100\% & \colorbox{green!30}{\textbf{48.7}} & 60.5 & 60.0\\
     ByteTrack~\cite{bytetrack} & \modelname & 3.3\% & 47.9 & 57.6 & 61.4\\
     ByteTrack~\cite{bytetrack} & \modelname & 10\% & 48.4 & \colorbox{green!30}{\textbf{61.0}} & \colorbox{green!30}{\textbf{62.8}}\\
     \midrule
     GHOST~\cite{ghost} & GT & 100\% & \colorbox{green!30}{\textbf{48.3}} & \colorbox{green!30}{\textbf{59.7}} & 60.2 \\
     GHOST~\cite{ghost} & \modelname & 3.3\%  & 47.0 & 58.2 & 60.7 \\
     GHOST~\cite{ghost} & \modelname & 10\%  & 46.9 & 59.0 & \colorbox{green!30}{\textbf{60.8}} \\

     \midrule
     \multicolumn{6}{c}{DanceTrack Test} \\
     \midrule
     ByteTrack~\cite{bytetrack} & GT & 100\% & 41.0 & 79.1 & 46.0\\
     ByteTrack~\cite{bytetrack} & \modelname & 3.3\% & 39.5 & 76.4 & 45.0\\
     ByteTrack~\cite{bytetrack} & \modelname & 20\% & \colorbox{green!30}{\textbf{41.3}} & \colorbox{green!30}{\textbf{80.9}} & \colorbox{green!30}{\textbf{48.1}}\\
    \midrule
    GHOST~\cite{ghost} & GT & 100\% & \colorbox{green!30}{\textbf{45.8}} & 80.6 & 46.3 \\
    GHOST~\cite{ghost} & \modelname & 3.3\%  & 41.0 & 76.3 & 44.8 \\
    GHOST~\cite{ghost} & \modelname & 20\%  & 43.9 & \colorbox{green!30}{\textbf{81.3}} & \colorbox{green!30}{\textbf{47.7}} \\
    \midrule

    \end{tabular}}

\caption{Evaluation of \modelname labels on MOT17, MOT20, and DanceTrack test sets with varying annotation budgets by training ByteTrack~\cite{bytetrack} and GHOST~\cite{ghost} using annotations generated by our method.}
\vspace{-0.5cm}
\label{table:benchmark_autolabels}
\end{table}

\PAR{How much data do we really need to label?} In this final experiment, our goal is to assess the impact of our entire pipeline by answering the question: How much data do we actually need to label in order to achieve the same performance as a model trained on the entire groundtruth training dataset?
For that, we choose two well-established trackers, ByteTrack~\cite{bytetrack} and GHOST~\cite{ghost}, and train them from scratch using our \modelname labels. We report their performance on MOT17, MOT20, and DanceTrack test sets in ~\cref{table:benchmark_autolabels}.
We first confirm that our results on the validation set also generalize to the test set, and that we can reach GT performance with $3.3\%$ of annotation effort on MOT17. Note that for MOT17 our annotations yield a slight improvement over GT, which can be attributed to observed annotation noise in the original GT, particularly in moving sequences (further details can be found in the supplementary material). 
MOT20 and DanceTrack are known to be more challenging datasets~\cite{dancetrack}, which is confirmed by the fact that the trackers still miss a few points when trained on only $3.3\%$ of manually annotated data.
Given that we cannot do a full analysis of different percentages because we are working on the test sets, we choose two percentages of manually annotated data on which to train our trackers: $10\%$ for MOT20, and $20\%$ for DanceTrack. In both cases, we achieve on-par performance with the trackers trained on full groundtruth data.
This clearly shows the potential in properly leveraging synthetic data, pseudo-labels from a strong model, and active learning on graph hierarchies in order to label multi-object tracking data, where we can save $96.7\%$ of the annotation effort on MOT17, $90\%$ of the annotation effort on MOT20 and $80\%$ of the annotation effort on DanceTrack.

\section{Conclusion}
\label{sec:conclusion}
We have introduced SPAM, a labeling engine to efficiently obtain trajectory annotations from videos. We have shown that SPAM is able to produce annotations that yield state-of-the-art tracking performance with a significantly reduced human annotation effort. Through our extensive experiments we have shown that (i) our synthetic pretraining + pseudo-label finetuning recipe yields a strong, annotation-free, labeler; (ii) our combination of active learning + pseudo-labeling for annotation generation yields high quality and efficient annotations; and (iii) that graphs are an excellent tool to provide comprehensive modelling for trajectory annotation. We believe that SPAM paves the way to make large-scale cheap annotation of tracking datasets a reality, which will in turn power the next generation of data-hungry tracking algorithms.

\par\vfill\par

\clearpage  

%
%
\bibliographystyle{splncs04}
\bibliography{egbib}

\begin{thebibliography}{10}
\providecommand{\url}[1]{\texttt{#1}}
\providecommand{\urlprefix}{URL }
\providecommand{\doi}[1]{https://doi.org/#1}

\bibitem{AcunaCVPR18}
Acuna, D., Ling, H., Kar, A., Fidler, S.: Efficient interactive annotation of segmentation datasets with polygon-rnn++. In: CVPR (2018)

\bibitem{Alahi_2016_CVPR}
Alahi, A., Goel, K., Ramanathan, V., Robicquet, A., Fei-Fei, L., Savarese, S.: Social lstm: Human trajectory prediction in crowded spaces. In: Proceedings of the IEEE Conference on Computer Vision and Pattern Recognition (CVPR) (June 2016)

\bibitem{5995311}
Andriyenko, A., Schindler, K.: Multi-target tracking by continuous energy minimization. In: CVPR. pp. 1265--1272 (2011). \doi{10.1109/CVPR.2011.5995311}

\bibitem{bastani2021self}
Bastani, F., He, S., Madden, S.: Self-supervised multi-object tracking with cross-input consistency. Advances in Neural Information Processing Systems  \textbf{34},  13695--13706 (2021)

\bibitem{ensemble}
Beluch, W.H., Genewein, T., N{\"u}rnberger, A., K{\"o}hler, J.M.: The power of ensembles for active learning in image classification. In: CVPR. pp. 9368--9377 (2018)

\bibitem{berclaz2011multiple}
Berclaz, J., Fleuret, F., Turetken, E., Fua, P.: Multiple object tracking using k-shortest paths optimization. IEEE TPAMI  \textbf{33}(9),  1806--1819 (2011)

\bibitem{tracktor}
Bergmann, P., Meinhardt, T., Leal-Taixe, L.: Tracking without bells and whistles. In: ICCV. pp. 941--951 (2019)

\bibitem{MPNTrackSeg}
Bras{\'o}, G., Cetintas, O., Leal-Taix{\'e}, L.: Multi-object tracking and segmentation via neural message passing. IJCV  \textbf{130}(12),  3035--3053 (2022)

\bibitem{mpntrack}
Braso, G., Leal-Taixe, L.: Learning a neural solver for multiple object tracking. In: CVPR (2020)

\bibitem{cai2022memot}
Cai, J., Xu, M., Li, W., Xiong, Y., Xia, W., Tu, Z., Soatto, S.: Memot: Multi-object tracking with memory. In: Proceedings of the IEEE/CVF Conference on Computer Vision and Pattern Recognition. pp. 8090--8100 (2022)

\bibitem{swav}
Caron, M., Misra, I., Mairal, J., Goyal, P., Bojanowski, P., Joulin, A.: Unsupervised learning of visual features by contrasting cluster assignments. NeurIPS  (2020)

\bibitem{sushi}
Cetintas, O., Bras\'o, G., Leal-Taix\'e, L.: Unifying short and long-term tracking with graph hierarchies. In: CVPR. pp. 22877--22887 (June 2023)

\bibitem{chen2018real}
Chen, L., Ai, H., Zhuang, Z., Shang, C.: Real-time multiple people tracking with deeply learned candidate selection and person re-identification. In: 2018 IEEE international conference on multimedia and expo (ICME). pp.~1--6. IEEE (2018)

\bibitem{simclr}
Chen, T., Kornblith, S., Norouzi, M., Hinton, G.: A simple framework for contrastive learning of visual representations. In: IEEE Int. Conf. Mach. Learn. pp. 1597--1607. PMLR (2020)

\bibitem{famnet}
Chu, P., Ling, H.: Famnet: Joint learning of feature, affinity and multi-dimensional assignment for online multiple object tracking. In: ICCV (October 2019)

\bibitem{dai2021video}
Dai, K., Zhao, J., Wang, L., Wang, D., Li, J., Lu, H., Qian, X., Yang, X.: Video annotation for visual tracking via selection and refinement. In: ICCV (2021)

\bibitem{up_detr}
Dai, Z., Cai, B., Lin, Y., Chen, J.: Up-detr: Unsupervised pre-training for object detection with transformers. In: Proceedings of the IEEE/CVF Conference on Computer Vision and Pattern Recognition (CVPR). pp. 1601--1610 (June 2021)

\bibitem{motchaijcv}
Dendorfer, P., Osep, A., Milan, A., Schindler, K., Cremers, D., Reid, I., Roth, S., Leal-Taix{\'e}, L.: Motchallenge: A benchmark for single-camera multiple target tracking. International Journal of Computer Vision  \textbf{129}(4),  845--881 (2021)

\bibitem{mot20}
Dendorfer, P., Rezatofighi, H., Milan, A., Shi, J.Q., Cremers, D., Reid, I.D., Roth, S., Schindler, K., Leal-Taix'e, L.: Mot20: A benchmark for multi object tracking in crowded scenes. ArXiv  \textbf{abs/2003.09003} (2020)

\bibitem{dollar2009pedestrian}
Doll{\'a}r, P., Wojek, C., Schiele, B., Perona, P.: Pedestrian detection: A benchmark. In: CVPR. pp. 304--311. IEEE (2009)

\bibitem{al_ili}
Elezi, I., Yu, Z., Anandkumar, A., Leal-Taixe, L., Alvarez, J.M.: Not all labels are equal: Rationalizing the labeling costs for training object detection. In: CVPR. pp. 14492--14501 (2022)

\bibitem{4587581}
Ess, A., Leibe, B., Schindler, K., Van~Gool, L.: A mobile vision system for robust multi-person tracking. In: CVPR. pp.~1--8 (2008). \doi{10.1109/CVPR.2008.4587581}

\bibitem{motsynth}
Fabbri, M., Bras\'o, G., Maugeri, G., Cetintas, O., Gasparini, R., O\v{s}ep, A., Calderara, S., Leal-Taix\'e, L., Cucchiara, R.: Motsynth: How can synthetic data help pedestrian detection and tracking? In: ICCV. pp. 10849--10859 (October 2021)

\bibitem{fabbri2018learning}
Fabbri, M., Lanzi, F., Calderara, S., Palazzi, A., Vezzani, R., Cucchiara, R.: Learning to detect and track visible and occluded body joints in a virtual world. In: ECCV (2018)

\bibitem{dttd}
Feichtenhofer, C., Pinz, A., Zisserman, A.: Detect to track and track to detect. In: ICCV (Oct 2017)

\bibitem{gaidon2016virtual}
Gaidon, A., Wang, Q., Cabon, Y., Vig, E.: Virtual worlds as proxy for multi-object tracking analysis. In: CVPR. pp. 4340--4349 (2016)

\bibitem{ge2021yolox}
Ge, Z., Liu, S., Wang, F., Li, Z., Sun, J.: Yolox: Exceeding yolo series in 2021. arXiv preprint arXiv:2107.08430  (2021)

\bibitem{he2016deep}
He, K., Zhang, X., Ren, S., Sun, J.: Deep residual learning for image recognition. In: Proceedings of the IEEE conference on computer vision and pattern recognition. pp. 770--778 (2016)

\bibitem{Hu3DT19}
Hu, H.N., Cai, Q.Z., Wang, D., Lin, J., Sun, M., Krähenbühl, P., Darrell, T., Yu, F.: Joint monocular 3d vehicle detection and tracking. In: ICCV (2019)

\bibitem{jabri2020space}
Jabri, A., Owens, A., Efros, A.: Space-time correspondence as a contrastive random walk. Advances in neural information processing systems  \textbf{33},  19545--19560 (2020)

\bibitem{4270205}
Jiang, H., Fels, S., Little, J.J.: A linear programming approach for multiple object tracking. In: CVPR. pp.~1--8 (2007). \doi{10.1109/CVPR.2007.383180}

\bibitem{clear}
Kasturi, R., Goldgof, D., Soundararajan, P., Manohar, V., Garofolo, J., Boonstra, M., Korzhova, V., Zhang, J.: Framework for performance evaluation for face, text and vehicle detection and tracking in video: data, metrics, and protocol. IEEE TPAMI  (2009)

\bibitem{Kaul22}
Kaul, P., Xie, W., Zisserman, A.: Label, verify, correct: A simple few-shot object detection method. In: IEEE Conference on Computer Vision and Pattern Recognition (2022)

\bibitem{adam}
Kingma, D.P., Ba, J.: Adam: A method for stochastic optimization. arXiv preprint arXiv:1412.6980  (2014)

\bibitem{kirillov2023segment}
Kirillov, A., Mintun, E., Ravi, N., Mao, H., Rolland, C., Gustafson, L., Xiao, T., Whitehead, S., Berg, A.C., Lo, W.Y., et~al.: Segment anything. In: ICCV (2023)

\bibitem{lan2023vision}
Lan, S., Yang, X., Yu, Z., Wu, Z., Alvarez, J.M., Anandkumar, A.: Vision transformers are good mask auto-labelers. In: IEEE Conference on Computer Vision and Pattern Recognition (CVPR) (June 2023)

\bibitem{Leal-Taixe_2016_CVPR_Workshops}
Leal-Taixe, L., Canton-Ferrer, C., Schindler, K.: Learning by tracking: Siamese cnn for robust target association. In: CVPRW (June 2016)

\bibitem{Leal-Taixe_2014_CVPR}
Leal-Taixe, L., Fenzi, M., Kuznetsova, A., Rosenhahn, B., Savarese, S.: Learning an image-based motion context for multiple people tracking. In: CVPR (June 2014)

\bibitem{6130233}
Leal-Taixé, L., Pons-Moll, G., Rosenhahn, B.: Everybody needs somebody: Modeling social and grouping behavior on a linear programming multiple people tracker. In: Int. Conf. Comput. Vis. Worksh. pp. 120--127 (2011). \doi{10.1109/ICCVW.2011.6130233}

\bibitem{hetero_AL_MOT_2023}
Li, R., Zhang, B., Liu, J., Liu, W., Zhao, J., Teng, Z.: Heterogeneous diversity driven active learning for multi-object tracking. In: ICCV (2023)

\bibitem{litrico_2023_CVPR}
Litrico, M., Del~Bue, A., Morerio, P.: Guiding pseudo-labels with uncertainty estimation for source-free unsupervised domain adaptation. In: Proceedings of the IEEE/CVF Conference on Computer Vision and Pattern Recognition (CVPR) (2023)

\bibitem{liu2023uncertainty}
Liu, K., Jin, S., Fu, Z., Chen, Z., Jiang, R., Ye, J.: Uncertainty-aware unsupervised multi-object tracking. In: Proceedings of the IEEE/CVF International Conference on Computer Vision. pp. 9996--10005 (2023)

\bibitem{hota}
Luiten, J., Osep, A., Dendorfer, P., Torr, P., Geiger, A., Leal-Taix{\'e}, L., Leibe, B.: Hota: A higher order metric for evaluating multi-object tracking. IJCV  \textbf{129}(2),  548--578 (2021)

\bibitem{manen2017pathtrack}
Manen, S., Gygli, M., Dai, D., Van~Gool, L.: Pathtrack: Fast trajectory annotation with path supervision. In: Proceedings of the IEEE International Conference on Computer Vision. pp. 290--299 (2017)

\bibitem{trackformer}
Meinhardt, T., Kirillov, A., Leal-Taixe, L., Feichtenhofer, C.: Trackformer: Multi-object tracking with transformers. In: IEEE Conf. Comput. Vis. Pattern Recog. (2022)

\bibitem{meng2023tracking}
Meng, S., Shao, D., Guo, J., Gao, S.: Tracking without label: Unsupervised multiple object tracking via contrastive similarity learning. In: Proceedings of the IEEE/CVF International Conference on Computer Vision. pp. 16264--16273 (2023)

\bibitem{motchallenge}
Milan, A., Leal-Taix{\'e}, L., Reid, I., Roth, S., Schindler, K.: Mot16: A benchmark for multi-object tracking. arXiv preprint arXiv:1603.00831  (2016)

\bibitem{qdtrack}
Pang, J., Qiu, L., Li, X., Chen, H., Li, Q., Darrell, T., Yu, F.: Quasi-dense similarity learning for multiple object tracking. In: CVPR. pp. 164--173 (2021)

\bibitem{5459260}
Pellegrini, S., Ess, A., Schindler, K., van Gool, L.: You'll never walk alone: Modeling social behavior for multi-target tracking. In: ICCV. pp. 261--268 (2009). \doi{10.1109/ICCV.2009.5459260}

\bibitem{5995604}
Pirsiavash, H., Ramanan, D., Fowlkes, C.C.: Globally-optimal greedy algorithms for tracking a variable number of objects. In: CVPR. pp. 1201--1208 (2011). \doi{10.1109/CVPR.2011.5995604}

\bibitem{motiontrack}
Qin, Z., Zhou, S., Wang, L., Duan, J., Hua, G., Tang, W.: Motiontrack: Learning robust short-term and long-term motions for multi-object tracking. In: Proceedings of the IEEE/CVF Conference on Computer Vision and Pattern Recognition. pp. 17939--17948 (2023)

\bibitem{IDF1}
Ristani, E., Solera, F., Zou, R., Cucchiara, R., Tomasi, C.: Performance measures and a data set for multi-target, multi-camera tracking. In: Eur. Conf. Comput. Vis. Worksh. pp. 17--35. Springer (2016)

\bibitem{Ristani_2018_CVPR}
Ristani, E., Tomasi, C.: Features for multi-target multi-camera tracking and re-identification. In: CVPR (June 2018)

\bibitem{Robicquet2016LearningSE}
Robicquet, A., Sadeghian, A., Alahi, A., Savarese, S.: Learning social etiquette: Human trajectory understanding in crowded scenes. In: ECCV (2016)

\bibitem{scovanner2009learning}
Scovanner, P., Tappen, M.F.: Learning pedestrian dynamics from the real world. In: 2009 IEEE 12th International Conference on Computer Vision. pp. 381--388. IEEE (2009)

\bibitem{ghost}
Seidenschwarz, J., Bras{\'o}, G., Serrano, V.C., Elezi, I., Leal-Taix{\'e}, L.: Simple cues lead to a strong multi-object tracker. In: CVPR. pp. 13813--13823 (2023)

\bibitem{coreset}
Sener, O., Savarese, S.: Active learning for convolutional neural networks: A core-set approach. In: ICLR (2018), \url{https://openreview.net/forum?id=H1aIuk-RW}

\bibitem{shao2018crowdhuman}
Shao, S., Zhao, Z., Li, B., Xiao, T., Yu, G., Zhang, X., Sun, J.: Crowdhuman: A benchmark for detecting human in a crowd. arXiv preprint arXiv:1805.00123  (2018)

\bibitem{dancetrack}
Sun, P., Cao, J., Jiang, Y., Yuan, Z., Bai, S., Kitani, K., Luo, P.: Dancetrack: Multi-object tracking in uniform appearance and diverse motion. In: CVPR (2022)

\bibitem{sun20eccv}
Sun, S., Akhtar, N., Song, X., Song, H., Mian, A., Shah, M.: Simultaneous detection and tracking with motion modelling for multiple object tracking. In: ECCV. pp. 626--643 (2020)

\bibitem{voigtlaender2019mots}
Voigtlaender, P., Krause, M., Osep, A., Luiten, J., Sekar, B.B.G., Geiger, A., Leibe, B.: Mots: Multi-object tracking and segmentation. In: Proceedings of the ieee/cvf conference on computer vision and pattern recognition. pp. 7942--7951 (2019)

\bibitem{voigtlaender2021reducing}
Voigtlaender, P., Luo, L., Yuan, C., Jiang, Y., Leibe, B.: Reducing the annotation effort for video object segmentation datasets. In: Proceedings of the IEEE/CVF Winter Conference on Applications of Computer Vision. pp. 3060--3069 (2021)

\bibitem{vondrick2011video}
Vondrick, C., Ramanan, D.: Video annotation and tracking with active learning. NeurIPS  \textbf{24} (2011)

\bibitem{vondrick2010efficiently}
Vondrick, C., Ramanan, D., Patterson, D.: Efficiently scaling up video annotation with crowdsourced marketplaces. In: Computer Vision--ECCV 2010: 11th European Conference on Computer Vision, Heraklion, Crete, Greece, September 5-11, 2010, Proceedings, Part IV 11. pp. 610--623. Springer (2010)

\bibitem{vondrick2018tracking}
Vondrick, C., Shrivastava, A., Fathi, A., Guadarrama, S., Murphy, K.: Tracking emerges by colorizing videos. In: Proceedings of the European conference on computer vision (ECCV). pp. 391--408 (2018)

\bibitem{CVPR2019_CycleTime}
Wang, X., Jabri, A., Efros, A.A.: Learning correspondence from the cycle-consistency of time. In: CVPR (2019)

\bibitem{9561110}
Wang, Y., Kitani, K., Weng, X.: Joint object detection and multi-object tracking with graph neural networks. In: IEEE Int. Conf. Robotics and Autom. pp. 13708--13715 (2021). \doi{10.1109/ICRA48506.2021.9561110}

\bibitem{jde}
Wang, Z., Zheng, L., Liu, Y., Wang, S.: Towards real-time multi-object tracking. The European Conference on Computer Vision (ECCV)  (2020)

\bibitem{wei2021aligning}
Wei, F., Gao, Y., Wu, Z., Hu, H., Lin, S.: Aligning pretraining for detection via object-level contrastive learning. In: Advances in Neural Information Processing Systems (2021)

\bibitem{xiao2017joint}
Xiao, T., Li, S., Wang, B., Lin, L., Wang, X.: Joint detection and identification feature learning for person search. In: CVPR. pp. 3415--3424 (2017)

\bibitem{5995468}
Yamaguchi, K., Berg, A.C., Ortiz, L.E., Berg, T.L.: Who are you with and where are you going? In: CVPR. pp. 1345--1352 (2011). \doi{10.1109/CVPR.2011.5995468}

\bibitem{utm}
You, S., Yao, H., Bao, B.K., Xu, C.: Utm: A unified multiple object tracking model with identity-aware feature enhancement. In: Proceedings of the IEEE/CVF Conference on Computer Vision and Pattern Recognition. pp. 21876--21886 (2023)

\bibitem{yuan2023active}
Yuan, D., Chang, X., Liu, Q., Yang, Y., Wang, D., Shu, M., He, Z., Shi, G.: Active learning for deep visual tracking. IEEE Transactions on Neural Networks and Learning Systems  (2023)

\bibitem{zeng2022motr}
Zeng, F., Dong, B., Zhang, Y., Wang, T., Zhang, X., Wei, Y.: Motr: End-to-end multiple-object tracking with transformer. In: European Conference on Computer Vision. pp. 659--675. Springer (2022)

\bibitem{network_flows_tracking}
Zhang, L., Li, Y., Nevatia, R.: Global data association for multi-object tracking using network flows. In: CVPR (2008)

\bibitem{zhang2017citypersons}
Zhang, S., Benenson, R., Schiele, B.: Citypersons: A diverse dataset for pedestrian detection. In: CVPR. pp. 3213--3221 (2017)

\bibitem{bytetrack}
Zhang, Y., Sun, P., Jiang, Y., Yu, D., Weng, F., Yuan, Z., Luo, P., Liu, W., Wang, X.: Bytetrack: Multi-object tracking by associating every detection box. In: ECCV. pp. 1--21. Springer (2022)

\bibitem{zhang2021fairmot}
Zhang, Y., Wang, C., Wang, X., Zeng, W., Liu, W.: Fairmot: On the fairness of detection and re-identification in multiple object tracking. IJCV  \textbf{129}(11),  3069--3087 (2021)

\bibitem{zheng2017person}
Zheng, L., Zhang, H., Sun, S., Chandraker, M., Yang, Y., Tian, Q.: Person re-identification in the wild. In: CVPR. pp. 1367--1376 (2017)

\bibitem{centertrack}
Zhou, X., Koltun, V., Kr{\"a}henb{\"u}hl, P.: Tracking objects as points. In: ECCV. pp. 474--490. Springer (2020)

\end{thebibliography}

\clearpage

\appendix

\begin{center}
    {\Large \bfseries Appendix}
\end{center}

\begin{abstract}
This supplementary material offers additional details (\cref{sec:more_spam}) and experimental results (\cref{sec:more_exps}) of \modelname.
To this end, we provide further information on the architecture of our graph-based model (\cref{sec:more_spam_model}) and its application for hierarchical labeling (\cref{sec:more_spam_active_learn_hier}).
In~\cref{sec:more_spam_implementation_details}, we complete the implementation details of the main paper.
The experimental results in \cref{sec:more_exps_frame} demonstrate the superiority of our approach to a common frame-based labeling method.
We compare labeling performance for static and moving cameras in \cref{sec:more_exps_label_noise_moving_cam} and finally discuss about an alternative evaluation setting in \cref{sec:more_exps_alternative_eval}.
\end{abstract}

\appendix
\section{Additional Details about \modelname}
\label{sec:more_spam}

\subsection{GNN \textcolor{\colorM}{\textbf{M}}odel }
\label{sec:more_spam_model}
Our hierarchical graph-neural network (GNN) utilizes the time-aware message passing framework introduced in~\cite{mpntrack}. We use a hierarchy of 10 GNNs operating over a video clip of 512 frames. At the first level, we employ a $GNN_{node}$, which considers a temporal window of 10 frames for each node. Subsequent levels, denoted as $GNN_{edge}$, cover 2, 4, ..., up to 512 frames. We follow a simple and unified approach for all hierarchy levels: $GNN_{node}$ and $GNN_{edge}$ levels share the same input features and network architecture (as presented in Sec 3.3 of the main paper). Moreover, $GNN_{edge}$ levels also share learnable weights across hierarchy levels. Overall, we rely on a lightweight GNN hierarchy with only approximately 65K parameters running at 20 frames per second (FPS).

\subsection{\textcolor{\colorA}{\textbf{A}}ctive Hierarchical Labeling}
\label{sec:more_spam_active_learn_hier}
To fully utilize our hierarchical graph framework, we distribute our annotation budget $B$ across $L$ hierarchical levels as $B_1, \dots, B_L$, with $B_1 + \dots + B_L = B$. Prioritizing deeper levels enables a more effective allocation of the annotation budget. Intuitively, in deeper hierarchy levels, nodes represent tracklets rather than individual detections. We propagate annotator decisions for entire clusters rather than individual detections, resolving multiple uncertainties with a single annotation. This makes the annotation pipeline more efficient compared to working at the detection level. For example for MOT17~\cite{motchallenge}, 50\% of the budget $B$ is allocated to the last three levels, 30\% to the previous three levels, and the remaining 20\% to earlier levels. For MOT17, we do not allocate a budget for refining bounding boxes since our experiments revealed that the bounding boxes generated by our synthetic detector are already well localized. However, for the challenging DanceTrack~\cite{dancetrack} dataset, we allocate a portion (approximately 30\%) of our annotation budget $B$ for refining bounding boxes.

\subsection{Implementation Details}
\label{sec:more_spam_implementation_details}

\PAR{Architecture and training.}
We train our YOLOX~\cite{ge2021yolox} detector on MOTSynth for 170 epochs following the training parameters in~\cite{bytetrack}. We keep augmentations on throughout the training. We train our GNN hierarchy on MOTSynth jointly for 10 epochs, with a learning rate of $3*10^{-4}$ and weight decay of $10^{-4}$ with batches of 4 graphs. We use a focal loss with $\gamma = 1$ and the Adam optimizer~\cite{adam}.

\PAR{Experimental setup.}
After obtaining labels with SPAM, we train two well-established online trackers ByteTrack~\cite{bytetrack} and GHOST~\cite{ghost} to evaluate our label quality in the main paper. Our main goal is to compare our label quality with ground truth label quality, therefore, we start trainings from scratch and only train on the dataset of interest (200 epochs for MOT17, 150 epochs for MOT20 and 100 epochs for DanceTrack). During testing, we follow their default parameters~\cite{bytetrack, ghost}. 

\section{Additional Experiments}
\label{sec:more_exps}

\begin{figure}[b]
  \centering
  \includegraphics[width=0.5\columnwidth]{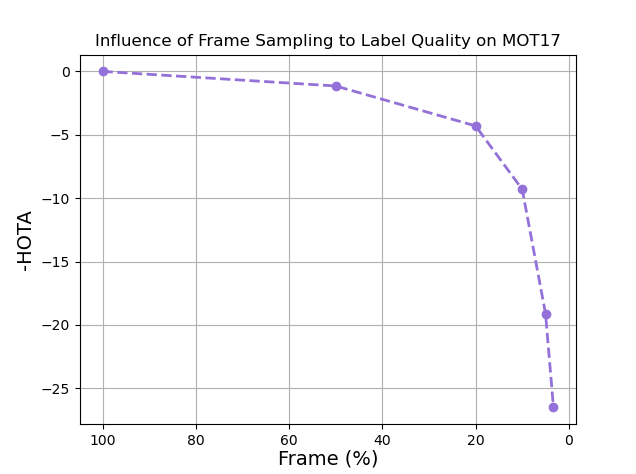}
  \caption{Annotation quality (reported as relative HOTA drop) on MOT17 by uniformly labeling different percentages of
frames, and then interpolating the ground truth boxes.}
  \label{fig:gt_interpolate}
\end{figure}

\subsection{Frame-based Labeling and FPS} 
\label{sec:more_exps_frame}

The most related work to active learning in the MOT domain is~\cite{hetero_AL_MOT_2023}, a frame-based labeling method. Motivated by the redundancy of nearby video frames, they propose to label only a portion of frames. 
However, their method requires a large labeling budget (50\% of the frames) to achieve ground-truth level performance. Furthermore, the authors~\cite{hetero_AL_MOT_2023} report that at lower data regimes (20\% to 40\%) their method is outperformed by a simple uniform sampling of frames. 

To demonstrate the inefficiency of such a frame-based labeling method, we show that for high frame rates, labeling such a large portion of the frames will trivially lead to ground-truth performance.
This is attributed to the redundancy of data in MOT17 sequences, which are annotated with high frames per second (FPS) of up to 30.
To this end, we examine the annotation quality on MOT17 by uniformly labeling different percentages of frames, \ie, reducing the FPS, and then interpolating the ground truth boxes. 
We report relative HOTA drops compared to labeling all frames in~\cref{fig:gt_interpolate}. Notably, we observe that a naive solution of labeling 50\% of the frames (every other frame) and interpolating the bounding boxes yields almost perfect labels (1 HOTA drop, leading to 99 HOTA). However, a significant performance drop is observed for lower data regimes, i.e., using lower than 20\% of frames equivalent to approximately 6 FPS for MOT17. This highlights that labeling becomes notably more challenging for low data regimes in which~\cite{hetero_AL_MOT_2023} fails to operate.

In contrast to frame-based labeling, \modelname performs active learning for multiple objects across all video frames, utilizing the uncertainty measures from our graph-based model. This allows us to invest expensive annotation efforts on \textit{individual tracks} instead of labeling \textit{entire frames} in a brute-force manner. Consequently, we can reach ground-truth level performance with a minimal annotation budget of only 3.3\% for MOT17.

\begin{figure}[tb]
  \centering
  \includegraphics[width=\columnwidth]{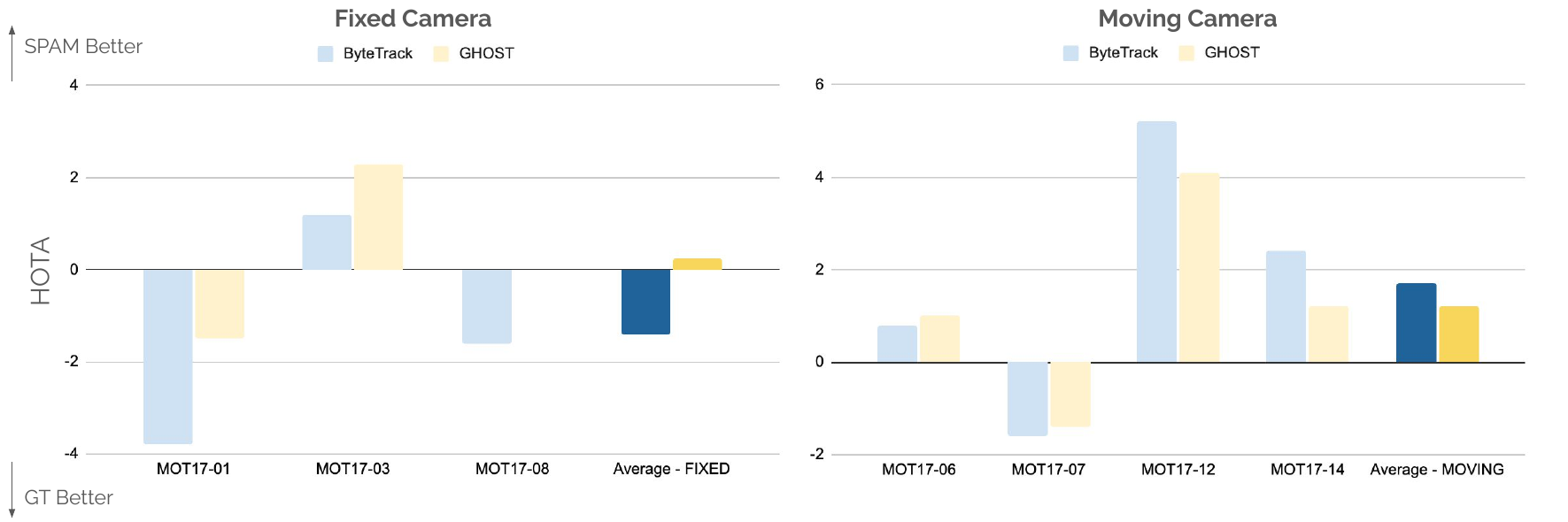}
  \caption{Comparison of MOT17 results for static and moving camera sequences.}
  \label{fig:seqs}
\end{figure}

\subsection{Label Noise for Moving Cameras}
\label{sec:more_exps_label_noise_moving_cam}

In our experiments detailed in Sec 4.6 of the main paper, we found that on MOT17, our annotations with a budget of 3.3\% yield slightly better performance compared to the ground truth. We attribute this observation to annotation noise in the original ground truth, particularly in sequences with moving cameras. To illustrate this, we provide detailed results on MOT17 for each sequence with ByteTrack and GHOST in~\cref{fig:seqs}.
We compare relative HOTA scores between trainings on ground truth and SPAM labels.
Positive scores indicate superior performance with SPAM labels, while negative scores indicate that ground truth labels result in better performance.
Overall, with a 3.3\% budget, ground truth exhibits better performance on static camera sequences, while SPAM outperforms ground truth labels on moving camera sequences.
We speculate that this phenomenon could be due to the additional noise in ground truth labels on moving sequences, which may be introduced by the interpolation-based labeling strategy reported in~\cite{motchaijcv}.

\subsection{Alternative Evaluation Setting}
\label{sec:more_exps_alternative_eval}
As demonstrated in the literature on pseudo-labeling, active learning, and noisy labels, labels do not contribute equally when training models. Therefore, our main goal is not to achieve perfect labels but to label data with minimal annotation effort while enabling optimal tracking performance. Thus, we train SOTA trackers with our labels and compare the results with ground truth training in our experiments similar to an AL evaluation setting. For completeness, we also report tracking scores reached by our labels directly as an alternative setting here. We obtain 84.1 MOTA, 72.2 HOTA, and 90.1 IDF1 with our labels (3.3\% annotation effort) on MOT17. This shows that the labels are indeed not completely covering groundtruth. However, results from the Table 5 of the main paper show that training on those labels gives the same results as training on full groundtruth. Thus, even though the labels are not covering the full ground-truth, allocating more annotation resources will not improve the training performance due to diminishing returns. This demonstrates that perfect labels are not necessary to reach optimal tracking performance and our experimental setting is well-suited for our goal.
\end{document}